\documentclass[letterpaper]{article} 
\usepackage{aaai23}  
\usepackage{times}  
\usepackage{helvet}  
\usepackage{courier}  
\usepackage[hyphens]{url}  
\usepackage{graphicx} 
\usepackage{natbib}  
\usepackage{caption} 
\usepackage{algorithm}
\usepackage{algorithmic}

\usepackage{enumitem}

\usepackage{newfloat}
\usepackage{listings}
\DeclareCaptionStyle{ruled}{labelfont=normalfont,labelsep=colon,strut=off} 
\floatstyle{ruled}
\newfloat{listing}{tb}{lst}{}
\floatname{listing}{Listing}
\title{Explaining (Sarcastic) Utterances to Enhance Affect Understanding in Multimodal Dialogues}
\author {
    Shivani Kumar \textsuperscript{\rm 1},
    Ishani Mondal \textsuperscript{\rm 2},
    Md Shad Akhtar \textsuperscript{\rm 1},
    Tanmoy Chakraborty \textsuperscript{\rm 3}
}
\affiliations {
    \textsuperscript{\rm 1}Indraprastha Institute of Information Technology Delhi, India\\
    \textsuperscript{\rm 2}University of Maryland, College Park\\
    \textsuperscript{\rm 3}Indian Institute of Technology Delhi, India\\
    shivaniku@iiitd.ac.in, ishani340@gmail.com, shad.akhtar@iiitd.ac.in, tanchak@iitd.ac.in
}

\usepackage{bibentry}

\usepackage{todonotes}
\usepackage{multirow}
\usepackage{graphicx}
\usepackage{amsfonts}
\usepackage{mathtools}
\usepackage{amsmath}

\usepackage{booktabs, multirow} 
\usepackage{soul}
\usepackage{xcolor,colortbl}

\definecolor{LightGreen}{rgb}{0.8,1,0.89}
\definecolor{LightRed}{rgb}{1.0,0.8,0.7}
\definecolor{LightCyan}{rgb}{0.88,1,1}

\usepackage{subfig}

\newcommand{\model}{\texttt{MOSES}}
\newcommand{\datasetSED}{\texttt{WITS}}
\newcommand{\datasetS}{\texttt{sWITS}}
\newcommand{\datasetH}{\texttt{hWITS}}
\newcommand{\datasetE}{\texttt{eWITS}}

\begin{document}

\maketitle

\begin{abstract}
Conversations emerge as the primary media for exchanging ideas and conceptions. From the listener's perspective, identifying various affective qualities, such as sarcasm, humour, and emotions, is paramount for comprehending the true connotation of the emitted utterance. However, one of the major hurdles faced in learning these affect dimensions is the presence of figurative language \textit{viz.} irony, metaphor, or sarcasm. We hypothesize that any detection system constituting the exhaustive and explicit presentation of the emitted utterance would improve the overall comprehension of the dialogue.     
To this end, we explore the task of Sarcasm Explanation in Dialogues that aims to unfold the hidden irony behind sarcastic utterances. We propose \model, a deep neural network, which takes a multimodal (sarcastic) dialogue instance as an input and generates a natural language sentence as its explanation. Subsequently, we leverage the generated explanation for various natural language understanding tasks in a conversational dialogue setup, such as \textit{sarcasm detection}, \textit{humour identification}, and \textit{emotion recognition}.
Our evaluation shows that \model\ outperforms the state-of-the-art system for SED by an average of $\sim2\%$ on different evaluation metrics, such as ROUGE, BLEU, and METEOR. Further, we observe that leveraging the generated explanation advances three downstream tasks for affect classification -- an average improvement of $\sim14\%$ F1-score in the sarcasm detection task and and $\sim2\%$ in the humour identification and emotion recognition task. We also perform extensive analyses to assess the quality of the results.  
\end{abstract}



\section{Introduction}
Expressing oneself eloquently to our conversation partner requires employing multiple affective components such as emotion, humour, and sarcasm.
All such attributes interact with each another to present a concrete definition of an uttered statement \citep{roberts1994people}.
While affects such as emotion and humour deem easier to comprehend, sarcasm, on the other hand, is a challenging aspect to comprehend \citep{olkoniemi2016individual}. 
Consequently, it becomes imperative for NLP systems to capture and understand sarcasm in its entirety. Sarcasm Explanation in Dialogues (SED) is a new task proposed recently in this direction \citep{kumar2022did}.
In this work, we scour the \textit{task of SED} and considerably improve the performance 
by proposing \model, a deep neural network which leverages the peculiarities of the benchmark dataset, \datasetSED.

\begin{figure}[t]
    \centering
    \includegraphics[width=\columnwidth]{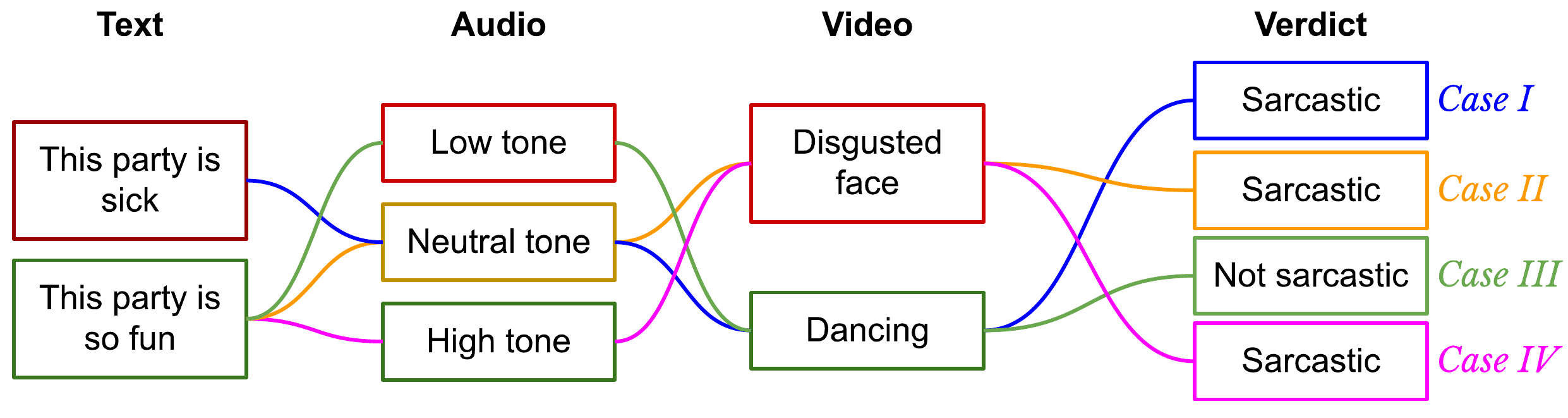}
    \caption{[Best viewed in color] Effect of multimodality on sarcasm. We do not show all possible combinations for brevity.}
    \label{fig:multimodal}
    \vspace{-5mm}
\end{figure}

Congruent to how humans make decisions after compiling data from all their available senses, (predominantly optical and auditory) multimodal analysis helps the machine mimic this behaviour. Particularly in the case of sarcasm interpretation, multimodal information provides us with essential indications to interpret irony. Figure \ref{fig:multimodal} shows cases where multimodal knowledge such as audio and video can assist in comprehending sarcasm. The identical text, ``This party is so fun", can result in diverse classes when integrated with different multimodal signals. For instance, when said with a \textit{neutral tone} and a \textit{disgusted face} (Case II), it results in a `sarcastic' verdict. On the other hand, when the same utterance is expressed with a \textit{low tone} and while the speaker is \textit{dancing} (Case III), we can assume the speaker means what they say without any sarcasm. To capture multimodality efficaciously, it is vital to grant a prime prerogative to each input modality in order to capture its' peculiarities. To this end, we propose a \textit{spotlight-aware fusion mechanism}, where the final multimodal amalgamated vector is tailored by paying special attention to individual modalities.

All affective components, such as sarcasm, humour, and emotion, work in tandem to convey a statement's intended meaning \citep{Hasan_Lee_Rahman_Zadeh_Mihalcea_Morency_Hoque_2021,chauhan-etal-2020-sentiment}. Accordingly, we hypothesize that understanding one of the affective markers, like sarcasm, in its entirety will influence comprehending others. Consequently, in this work, we deal with \textit{leveraging sarcasm explanations} for three affect understanding tasks in dialogues, namely sarcasm detection, humour identification, and emotion recognition.
The performance obtained from these tasks can be employed as a method to estimate the relevance of the SED task extrinsically.
We summarize our contributions below:
\begin{enumerate}[leftmargin=*,noitemsep,topsep=0pt]
    \item We explore the new task of  SED and propose {a novel model, \model}, for it.
    \item We compare \model\ with existing baselines and obtain {state-of-the-art results for the SED task}.
    \item We show the {application of the generated explanations} in understanding different affective components -- sarcasm, emotions and humour.
    \item We show {extensive quantitative and qualitative studies} for all our experiments.
\end{enumerate}

\textbf{Reproducibility:} The source code for \model\ with the execution instructions are present here: \url{https://github.com/LCS2-IIITD/MOSES.git}.

\section{Related Work}
\textbf{Sarcasm.} Figurative language such as sarcasm plays an integral role in resolving the veiled meaning of an uttered statement. Earlier studies dealt with sarcasm identification in standalone texts like tweets and reviews. \cite{kreuz-caucci:2007:sarcasm:lexical, tsur:sarcasm:2010, joshi:sarcsam:incongruity:2015, peled-reichart-2017-sarcasm}. A detailed summary of these studies can be found in the survey compiled by \citet{1joshi2017automatic}. 
Several work explored sarcasm in other languages such as Hindi \citep{bharti2017harnessing}, Arabic \citep{abu-farha-magdy-2020-arabic}, Spanish \citep{ortega2019overview}, Italian \citep{cignarella2018overview}, or even code-mixed \citep{swami2018corpus} languages.

\textbf{Sarcasm and Dialogues.} Linguistic and lexical traits were the primary sources of sarcasm markers in previous investigations \citep{kreuz-caucci:2007:sarcasm:lexical,tsur:sarcasm:2010}. However, in more contemporary studies, attention-based approaches are used to capture the inter- and intra-sentence interactions in the text \citep{tay-etal-2018-reasoning, xiong2019sarcasm, srivastava-etal-2020-novel}. In terms of conversations, \citet{ghosh2017role} harnessed attention-based RNNs to capture context and determinate sarcasm.

\textbf{Sarcasm and Multimodality.} \citet{castro2019multimodal} proposed a multimodal, multiparty, English dataset called MUStARD to benchmark the task of multimodal sarcasm identification in conversation. Subsequently, \citet{chauhan-etal-2020-sentiment} devised a multi-task framework by leveraging interdependency between emotions and sarcasm to solve the task of multimodal sarcasm detection. Another work \cite{Hasan_Lee_Rahman_Zadeh_Mihalcea_Morency_Hoque_2021} established the interdependency of humour with sarcasm by suggesting a humour knowledge enriched Transformer model for sarcasm detection. In the code-mixed scenario, \citet{9442359} proposed \textsc{MaSaC}, a multimodal, multiparty, code-mixed dialogue dataset for humour and sarcasm detection.
In the bimodal setting, sarcasm identification with tweets containing images has also been well explored \citep{cai-etal-2019-multi, xu-etal-2020-reasoning, pan-etal-2020-modeling} .

\textbf{Beyond Sarcasm Detection.}Sarcasm generation is another direction that practitioners are inquisitive about due to its forthright benefit in enhancing chatbot engagement. Thereby, \citet{mishra-etal-2019-modular} induced sarcastic utterances by presenting context incongruity through fact removal and incongruous phrase insertion. A retrieve-and-edit-based unsupervised approach for generating sarcasm was proposed by \citet{chakrabarty-etal-2020-r} that exploits semantic incongruity and valence reversal to convert non-sarcastic instances to sarcastic ones.
On the other hand, while detecting irony is crucial, it is insufficient to capture the cardinal connotation of the statement. Consequently, \citet{dubey2019deep} examined the task of converting sarcastic utterances into their non-sarcastic counterparts using deep neural networks.

In this work, we explore the task of Sarcasm Explanation in Dialogues, the second attempt after  \citet{kumar2022did}. SED aims to generate natural language explanations for a disseminated multimodal sarcastic conversation. We present a new model, \model, which enhances the current state-of-the-art for the SED task. However, unlike \citet{kumar2022did}, we perform both intrinsic and extrinsic evaluations to show the efficacy and usefulness of our model. We leverage the generated explanations to improve three affect understanding tasks -- sarcasm detection, humour identification, and emotion recognition in dialogues.

\section{Dataset}
Human conversations often take place employing a variety of languages. The phenomenon of using a blend of more than one language to communicate is dubbed code-mixing. Due to the prevalence of code-mixing in today's world, we consider the \datasetSED\ dataset \citep{kumar2022did}, which contains code-mixed dialogues (English-Hindi) from an Indian TV series. The dataset comprises multimodal, multiparty, code-mixed, sarcastic conversations where each sarcastic instance is annotated with a corresponding natural language code-mixed explanation.

In order to gauge the effect of sarcasm explanation on affective attributes, we augment the \datasetSED\ dataset to perform sarcasm detection, humour identification, and emotion recognition on it. We create instances for sarcastic and non-sarcastic utterances with their context to perform sarcasm detection. We call this variation of the dataset  \datasetS. Adapted from \textsc{MaSaC} \citep{9442359}, \datasetSED\ can also be mapped to annotations for humour identification, where each utterance contains a binary marker showcasing whether the utterance is amusing or not. Consequently, we map each instance in \datasetS\ to its corresponding humour annotation.
Additionally, we determine emotion labels for the instances at hand and identify the following emotions -- \textit{sadness, joy, anger}, {\em and neutral}. Three annotators were involved in this phase and achieved an inter-annotator agreement of $0.86$. More information can be found in the supplementary. Accordingly, we obtain four variations of the dataset:
\begin{enumerate}
    \item \datasetSED: It contains multimodal, multiparty, code-mixed, sarcastic instances with associated explanations.
    \item \datasetS: It contains sarcastic and non-sarcastic instances constructed from \datasetSED. The last utterance of each instance is marked by a binary tag indicating whether the statement contains sarcasm or not.
    \item \datasetH: For each instance created in \datasetS, each target utterance is marked with another binary label revealing the existence of humour in it.
    \item \datasetE: Similar to \datasetH, this variant contains emotion labels for the target utterances.
\end{enumerate}
Table \ref{tab:data-stats} illustrates the elementary statistics for the explained dataset variations. More details about the dataset are present in the supplementary.

\begin{table}[!t]\centering
\resizebox{\columnwidth}{!}{
\begin{tabular}{l|c|cc|cc|ccccc}\toprule
\multirow{2}{*}{} &\textbf{\datasetSED} &\multicolumn{2}{c|}{\textbf{\datasetS}} &\multicolumn{2}{c|}{\textbf{\datasetH}} &\multicolumn{4}{c}{\textbf{\datasetE}} \\\cmidrule{2-10}
&\textbf{\#S} &\textbf{\#NS} &\textbf{\#S} &\textbf{\#NH} &\textbf{\#H} &\textbf{\#Neutral} &\textbf{\#Sadness} &\textbf{\#Joy} &\textbf{\#Anger} \\\midrule
\textbf{Train} & $1792$ & $1669$ & $1792$ & $2795$ & $995$ & $1590$ & $1147$ & $623$ & $429$ \\
\textbf{Val} & $224$ & $213$ & $224$ & $362$ & $112$ & $196$ & $133$ & $87$ & $57$ \\
\textbf{Test} & $224$ & $218$ & $224$ & $367$ & $106$ & $195$ & $141$ & $70$ & $67$ \\\midrule
\textbf{Total} & $2240$ & $2100$ & $2240$ & $3524$ & $1213$ & $1981$ & $1421$ & $780$ & $553$ \\
\bottomrule
\end{tabular}}
\caption{Statistics of the sarcasm, humour, and emotion datasets in consideration (number of dialogue instances marked as sarcastic (\#S), non-sarcastic (\#NS), non-humorous (\#NH), and humorous (\#H).).}
\label{tab:data-stats}
\vspace{-5mm}
\end{table}

\section{Proposed Method}
\begin{figure*}[ht]
    \centering
    \includegraphics[width=\textwidth]{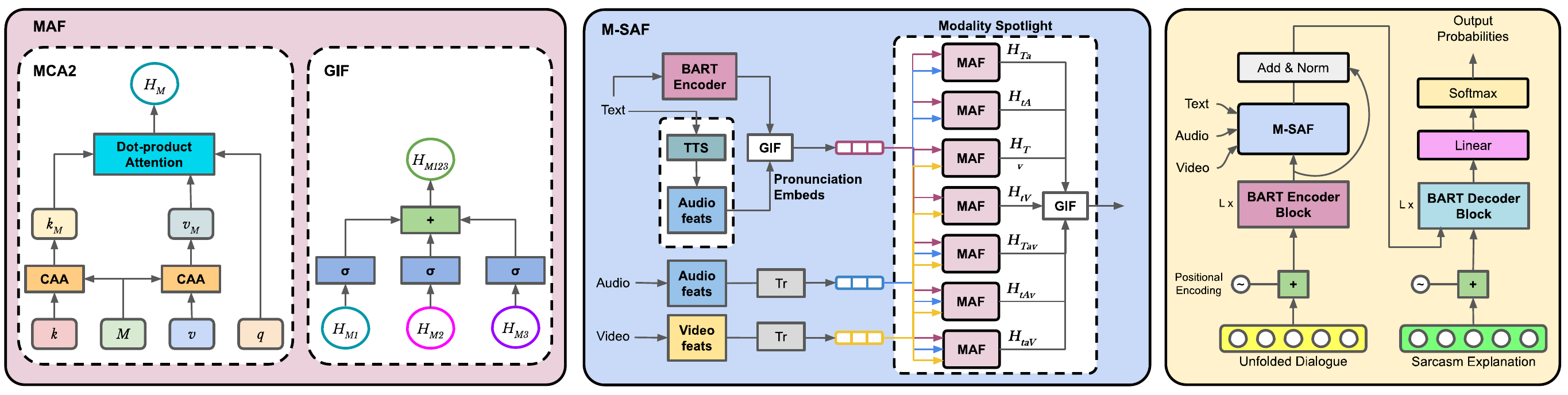}
    \caption{Model architecture for \model. The \textsc{MAF} model captures acoustic and visual hints using the Multimodal Context Aware Attention (MCA2) and combines them using Global Information Fusion (GIF). Each modality is kept in spotlight using the Modality Spotlight (MS) module. To capture the subjectivity in the code-mixed spellings, we propose to use  \textit{pronunciation embeddings}.}
    \label{fig:model}
    \vspace{-5mm}
\end{figure*}

This section illustrates the working of our proposed model, \model\ as presented in Figure \ref{fig:model}. The existing SED model, \textsc{MAF} \citep{kumar2022did}, which uses a modified version of context-aware attention \citep{Yang_Li_Wong_Chao_Wang_Tu_2019}, takes the multimodal (audio/video) vectors as context and fuses them with the text modality to generate multimodal fused text vectors. This way of multimodal fusion makes text the primary modality and treats the other signals (acoustic and visual) as secondary. Such a fusion technique might result in the downplay of the audio and video modalities. However, in the complete duration of the discourse, modalities other than text could play the deciding role in resolving the affects in consideration. Consequently, we propose using context-aware fusion in such a way that each modality gets a chance to play a pivotal role in the fusion computation.

The existing \textsc{MAF} module consists of an adapter-based module comprising two modules. The two modules -- Multimodal Context Aware Attention (MCA2) and Global Information Fusion (GIF) together make up the Multimodal Aware Fusion (MAF) module. Given the three input signals, namely text, audio, and video, the MCA2 module effectively introduces multimodal information in the textual representations. Further, the GIF module combines the multimodal infused textual representations. We insert another module in the pipeline, Modality Spotlight (MS), which is responsible for attending to each modality by treating it as the primary modality and the rest as the context. We explain each of these modules below. 

\subsubsection{Multimodal Context-Aware Attention (MCA2).}
The textual modality directly interacts with the other modalities in the standard fusion scheme, which uses dot-product based cross-modal attention. The multimodal representation acts as the key and value vectors while the text serves as the query. However, such a direct fusion of multimodal information from different embedding subspaces can lead to an inefficient representation that cannot capture contextual information. Consequently, inspired by \citet{Yang_Li_Wong_Chao_Wang_Tu_2019}, a context-aware attention block is used instead of dot product attention in the MCA2 module.

The multimodal context-aware attention first generates multimodal information conditioned key and value vectors using a primary representation, $H$ which depends on the choice of dominant modality in consideration. For example, $H$ can be obtained from any language model, such as BART \citep{lewis2019bart}, if text is to be considered as the primary modality. To generate information fused vectors, the MCA2 module first needs to convert $H$ into key, query, and value vectors, $q$, $k$, $v \in \mathbb{R}^{n\times{d}}$, respectively, as illustrated in Equation \ref{eq:equation1}. Here, $W_q$, $W_k$, and $W_v \in \mathbb{R}^{d\times{d}}$ are learnable parameters where the input, of maximum length $n$, is represented by a vector of dimension $d$.
\begin{equation}
    \label{eq:equation1}
    [qkv] = H[W_qW_kW_v]
\end{equation}
If we consider $M \in \mathbb{R}^{n\times{d_c}}$ a multimodal vector, the multimodal information infused key and value vectors $k_m$, $v_m \in \mathbb{R}^{n\times{d}}$, are generated following Equation \ref{eq:equation2}. Here, $\lambda \in \mathbb{R}^{n\times{1}}$ decides the amount of multimodal information to capture into the primary modality.
\begin{eqnarray}
    \label{eq:equation2}
    \begin{bmatrix}
    k_m \\ v_m
    \end{bmatrix} = (1 - \begin{bmatrix}
    \lambda_k \\ \lambda_v
    \end{bmatrix})\begin{bmatrix}
    k \\ v
    \end{bmatrix} + \begin{bmatrix}
    \lambda_k \\ \lambda_v
    \end{bmatrix}(M\begin{bmatrix}
    U_k \\ U_v
    \end{bmatrix})
\end{eqnarray}
The parameter $\lambda$ is learned using a gating mechanism, as shown in Equation \ref{eq:equation3}. Note that the matrices, $U_k$, $U_v \in \mathbb{R}^{d_c\times{d}}$, and $W_{k1}$, $W_{v1}$, $W_{k1}$, $W_{v2} \in \mathbb{R}^{d\times{1}}$, in Equations \ref{eq:equation2} and \ref{eq:equation3} are trained with the model.
\begin{eqnarray}
    \label{eq:equation3}
    \begin{bmatrix}
    \lambda_k \\ \lambda_v
    \end{bmatrix} = \sigma(\begin{bmatrix}
    k \\ v
    \end{bmatrix} \begin{bmatrix}
    W_{k_1} \\ W_{v_1}
    \end{bmatrix} + M\begin{bmatrix}
    U_k \\ U_v
    \end{bmatrix} \begin{bmatrix}
    W_{k_2} \\ W_{v_2}
    \end{bmatrix})
    \label{eq:lambda}
\end{eqnarray}
Once the modified key and value pair is obtained, the traditional dot-product attention is used to obtain the final multimodal fused vectors.

\subsubsection{Modality Spotlight (MS).}
We discussed how we can generate multimodal infused vector representation considering one modality as primary and the rest as context. Our work deals with three modalities -- text, acoustic, and visual. The spotlight module is responsible for treating each of these modalities as the primary modality at a time and generating the corresponding fused vectors. For instance, if text is considered the primary modality, then we need to calculate two multimodal fused vectors, $H_{Ta}$ and $H_{Tv}$, such as audio and video, play the role of context in the representations, respectively. Similarly, when audio and video are considered the primary source of information, $H_{tA}$ and $H_{tV}$ are calculated. Note that we do not calculate $H_{Av}$ or $H_{aV}$ because we are dealing with a textual generation task where the textual information plays the preliminary role.

Apart from bi-modal interactions, we also deal with tri-modal interactions in our work, where all three modalities are infused using the GIF module, as discussed later. Unlike bi-modal fusion, it is unfair to let text be the only primary modality in the tri-modal fusion. Consequently, we compute three tri-modal vectors, $H_{Tav}$, $H_{tAv}$, and $H_{taV}$, such that text, audio, and video individually play the primary role, respectively.

\subsubsection{Global Information Fusion (GIF).}
\label{sec:gif}
The GIF module is responsible for combining the information from multiple modalities together in an efficient manner. $G$ gates are used to control the amount of information disseminated by each modality, where $2 \leq G \leq 3$ is the number of modalities to fuse. For instance, if we calculate the interaction between the text and audio modalities with text being the primary source of information, we will first need to calculate the gated information from the audio representation using Equation \ref{eq:equation5}.
\begin{eqnarray}
    \label{eq:equation5}
    g_a = [H \oplus H_a] W_a + b_a 
\end{eqnarray}
where $W_a$ and $b_a$ are learnable matrices, and $\oplus$ denotes vector concatenation. The final representation to be passed on to the next encoder layer, in this case, will be obtained using Equation \ref{eq:equation6}.
\begin{eqnarray}
    \label{eq:equation6}
    H_{Ta} = H + g_a \odot H_a
    \label{eq:GIF}
\end{eqnarray}
On similar lines, if we are to calculate the tri-modal representation keeping the text as the primary modality, we first compute the gated vector for audio and video and then compute a weighted combination of the three modalities. The following sequence of equations illustrates this process,
\begin{equation*}
\begin{split}
    g_a &= [H \oplus H_a] W_a + b_a \nonumber \\
    g_v &= [H \oplus H_v] W_v + b_v \nonumber \\
    H_{Tav} &= H + g_a \odot H_a + g_v \odot H_v \nonumber
\end{split}
\end{equation*}
Likewise, we calculate the following set of vectors: $H_{Ta}$, $H_{tA}$, $H_{Tv}$, $H_{tV}$, $H_{Tav}$, $H_{tAv}$, and $H_{taV}$. Further, another GIF module is used to conglomerate these seven vectors, as shown in Equation \ref{eq:equation7}.
\begin{eqnarray}
    \label{eq:equation7}
    H_{all} = g_t \odot H + g_{Ta} \odot H_{Ta} + g_{tA} \odot H_{tA} + \nonumber \\
    g_{Tv} \odot H_{Tv} + g_{tV} \odot H_{tV} + g_{Tav} \odot H_{Tav} + \nonumber \\
    g_{tAv} \odot H_{tAv} + g_{taV} \odot H_{taV}
\end{eqnarray}

\section{Experiment and Results}
This section illustrates the feature extraction strategy we use and the baseline systems to which we compare our model, followed by the results we obtain for the SED task. We use the standard generative metrics -- ROUGE-1/2/L \citep{lin-2004-rouge}, BLEU-1/2/3/4 \citep{papineni2002bleu}, and METEOR \citep{denkowski:lavie:meteor-wmt:2014} to capture the syntactic and semantic performance of our system. Details about the execution process and the hyperparameters used are mentioned in the supplementary.

\subsection{Feature Extraction}
The primary challenges for generating vector representations for the instances in \datasetSED\ come from the code-mixed and multimodal aspects of the dataset. We alleviate these by proposing intelligent feature extraction methods.

\paragraph{Text:}
The textual input in \datasetSED\ is present in romanised code-mixed format.
Thereby, it may contain terms with the same meaning but varying spellings that are phonetically identical. For instance, the word \textit{``main"} in Hindi (translating to \textit{``I"} in English) can be written as \textit{``main"} or \textit{``mein"}.
To capture the similarity between all these spelling variations, we propose using \textit{Pronunciation Embeddings (PE)} that capture the phonetic equivalence between the words of the input text. We convert the text into a standard speech format using python's gTTS library\footnote{\url{https://pypi.org/project/gTTS/}}. This converted audio does not contain any tone or pitch variation for any term and thus, sounds the same for phonetically similar terms. We then extract the audio features from this converted speech using the method described below. This pronunciation vector is fused with the text representation, obtained from any encoder model like BART, using the GIF module to obtain the final text representation.

\paragraph{Audio:}
We use audio features provided by the authors of \datasetSED. They are $154$ dimension features capturing the loudness, and Mel Frequency Cepstral Coefficient (MFCCs) for each instance using the eGeMAPS model \cite{eyben2015geneva}. 
A Transformer encoder \cite{vaswani2017attention} is used for further processing of these features.

\paragraph{Video:}
Likewise, for the visual representation, we use the extracted features provided along with \datasetSED. These features are obtained using ResNext-$101$ \cite{hara2018spatiotemporal}, a pre-trained action recognition model that recognizes $101$ different actions.
A resolution of $720$ pixels, a window length of $16$, and a frame rate of $1.5$ are used to obtain $2048$ dimensional visual features. Analogous to the acoustic representation, a Transformer encoder captures the sequential conversation context in the resultant vectors.


\subsection{Comparative Systems} We use various established sequence-to-sequence (seq2seq) models to obtain the most promising textual representations for the discourse. \textbf{RNN}: The openNMT4\footnote{\url{https://github.com/OpenNMT/OpenNMT-py}} implementation of the RNN seq2seq architecture is used to obtain the results.
\textbf{Transformer} \citep{vaswani2017attention}: Explanations are generated using the vanilla Transformer encoder-decoder model.
\textbf{Pointer Generator Network (PGN)} \citep{see2017get}: A combination of generation and copying mechanisms is used in this seq2seq architecture.
\textbf{BART} \citep{lewis2019bart}: We use the base version of this denoising autoencoder model. It has a bidirectional encoder with an auto-regressive left-to-right decoder built on standard machine translation architecture.
\textbf{mBART} \citep{liu2020multilingual}: Trained on multiple large-scale monolingual corpora, mBART follows the same objective and architecture as BART\footnote{\url{https://huggingface.co/facebook/mbart-large-50-many-to-many-mmt}}.

\subsection{Results}
\paragraph{Textual:} Table \ref{tab:results} shows the results obtained when textual systems are used to obtain the generated explanations. We notice that while PGN delivers us with the least performance across most metrics, BART-based representations outperform the rest by providing the best performance across the majority of all evaluation metrics.

\begin{table}[!htp]\centering
\resizebox{\columnwidth}{!}{%
\begin{tabular}{l|c|ccccccccc}\toprule
\textbf{Mode} &\textbf{Model} &\textbf{R1} &\textbf{R2} &\textbf{RL} &\textbf{B1} &\textbf{B2} &\textbf{B3} &\textbf{B4} &\textbf{M} \\\cmidrule{1-10}
\multirow{5}{*}{\rotatebox{90}{\textbf{Textual}}} &\textbf{RNN} &29.22 &7.85 &27.59 &22.06 &8.22 &4.76 &2.88 &18.45 \\
&\textbf{Transformer} &29.17 &6.35 &27.97 &17.79 &5.63 &2.61 &0.88 &15.65 \\
&\textbf{PGN} &23.37 &4.83 &17.46 &17.32 &6.68 &1.58 &0.52 &23.54 \\
&\textbf{mBART} &33.66 &11.02 &31.5 &22.92 &10.56 &6.07 &3.39 &21.03 \\
&\textbf{BART} &36.88 &11.91 &33.49 &27.44 &12.23 &5.96 &2.89 &26.65 \\ \cmidrule{1-10}
\multirow{6}{*}{\rotatebox{90}{\textbf{Multimodality}}} &\textbf{\textsc{MAF-TA}} &38.21 &14.53 &35.97 &30.58 &15.36 &9.63 &5.96 &27.71 \\
&\textbf{\textsc{MAF-TV}} &37.48 &15.38 &35.64 &30.28 &16.89 &10.33 &6.55 &28.24 \\
&\textbf{\textsc{MAF-TAV}} &39.69 &17.1 &37.37 &33.2 &18.69 &12.37 &8.58 &30.4 \\\cmidrule{2-10}
&\textbf{\model\texttt{-TA}} &38.27 &14.53 &35.72 &31.57 &16.37 &9.66 &6.06 &29.27 \\
&\textbf{\model\texttt{-TV}} &39.62 &16.78 &37.48 &32.69 &17.76 &11.01 &6.89 &31.65 \\
&\textbf{\model\texttt{-TAV}} &40.88 &18.33 &38.38 &33.27 &18.87 &12.6 &8.8 &31.41
 \\\cmidrule{2-10}
&\textbf{\model} & \textbf{42.17} & \textbf{20.38} & \textbf{39.66} & \textbf{34.95} & \textbf{21.47} & \textbf{15.47} & \textbf{11.45} & \textbf{32.37}
 \\\midrule
\end{tabular}}
\caption{Experimental results (Abbreviation: R1/2/L: ROUGE1/2/L; B1/2/3/4: BLEU1/2/3/4; M: METEOR; PGN: Pointer Generator Network). Final row denotes \model\ including the pronunciation and spotlight modules.}
\label{tab:results}
\vspace{-4mm}
\end{table}

\paragraph{Pronunciation Embeddings (PE):} Due to the subjective nature of how other languages (Hindi, in our case) are written in a romanised format, the spellings of the words come from their phonetic understanding. To resolve the ambiguity between the same words with differing spellings, we propose to use pronunciation embeddings. As illustrated in Table \ref{tab:results}, we observe that by adding the PE component to the model with the help of the GIF module, the performance of text-based systems jumps by an average of $\sim4\%$ across all evaluation metrics.

\paragraph{Multimodality:} After we obtain the representation for the code-mixed text by fusing textual representation with pronunciation embeddings, we move on to adding multimodality to the system. We experimented with an established SED method (\textsc{MAF-TAV}) to estimate the effect of multimodality. Table \ref{tab:results} exhibits that while the addition of acoustic signals does not result in a performance gain, the addition of visual cues boosts the performance by $\sim1\%$ across all metrics. This phenomenon can be attributed to the fact that audio alone may cause confusion while understanding sarcasm, and visual hints may help in such times (Case III in Figure \ref{fig:multimodal}).
Thereby, improving the visual feature representations can be one of the future directions. Finally, when we add all multimodal signals together, we observe the best performance yet with an average increase of further $\sim1\%$ across majority metrics.

\paragraph{Modality Spotlight:} As hypothesised, we obtain the best performance for sarcasm understanding when all the three modalities are used in tandem. However, earlier methods for SED provided limelight to only textual representations \cite{kumar2022did}. We argue that especially in the case of sarcasm, multimodal signals such as audio and video might play the principal role in many instances. To comprehend this rotating importance of modalities, we use the spotlight module that aims to treat each modality as the primary modality while calculating the final representation. We observe an increase of $\sim2\%$ across all evaluation metrics as shown in Table \ref{tab:results}. These results directly support our hypothesis of the effect of multimodality in sarcasm analysis.

\begin{table}[!htp]\centering
\resizebox{\columnwidth}{!}{
\begin{tabular}{l|ccccccccc}\toprule
\textbf{Model} &\textbf{R1} &\textbf{R2} &\textbf{RL} &\textbf{B1} &\textbf{B2} &\textbf{B3} &\textbf{B4} &\textbf{M} \\\midrule
\textbf{BART} &36.88 &11.91 &33.49 &27.44 &12.23 &5.96 &2.89 &26.65 \\
\quad \textbf{+concat} &17.22 &1.7 &14.12 &13.11 &2.11 &0.0 &0.0 &9.34 \\
\quad \textbf{+DPA} &36.43 &13.04 &33.75 &28.73 &14.02 &8.0 &4.89 &25.6 \\
\quad \textbf{+MCA2} &36.37 &13.85 &34.92 &28.49 &14.34 &9.0 &6.16 &25.75 \\
\quad \quad \textbf{ + GIF} &39.69 &17.1 &37.37 &33.2 &18.69 &12.37 &8.58 &30.4 \\
\quad \quad \quad \textbf{ + PE} &40.88 &18.33 &38.38 &33.27 &18.87 &12.6 &8.8 &31.41 \\\midrule
\quad \quad \quad \quad \textbf{ + MS} (\model) &\textbf{42.17} &\textbf{20.38} &\textbf{39.66} &\textbf{34.95} &\textbf{21.47} &\textbf{15.47} &\textbf{11.45} &\textbf{32.37} \\
\bottomrule
\end{tabular}}
\caption{Ablation results on \model\ (DPA: Dot Product Attention).}
\label{tab:ablation}
\vspace{-5mm}
\end{table}
\begin{table*}[!htp]\centering
\resizebox{\textwidth}{!}{
\begin{tabular}{p{28em}|p{18em}p{18em}p{18em}p{18em}}\toprule
\textbf{Dialogue} &\textbf{Ground Truth} &\textbf{\textsc{MAF}} &\textbf{\model} \\\midrule
KISMI: Bas na Sahil bhai, meri firki kheech rahe ho na!? {\color{blue}\textit{(Enough brother Sahil, are you teasing me?!)}} & \multirow{2}{18em}{Sahil Kismi ko taunt maarta hai kyuki use rail gaadi ki awaaj sunni hai. {\color{blue}\textit{(Sahil taunts Kismi that she wants to hear the sound of a train)}}} & \multirow{2}{18em}{Sahil Kismi ko taunt maarta hai ki use pasand nahi. {\color{blue}\textit{(Sahil taunts Kismi that he doesn't like)}}} & \multirow{2}{18em}{Sahil Kismi ko taunt maarta hai kyuki use rail gaadi ki awaaj sunni hai. {\color{blue}\textit{(Sahil taunts Kismi that she wants to hear the sound of a train)}}} \\
SAHIL: Nahi, nahi, kya hai ki, mere CD ki collection mein na, ye train ke awaaj vali CD nahi hai... {\color{blue}\textit{(No no, see I don't have train's sound in my CD collection...)}} & & & \\ \midrule

MADHUSUDHAN: Kitne saal ka ho jaaega vo? {\color{blue}\textit{(How old will he be?)}} & \multirow{2}{18em}{Indravardhan Madhusudan ke questions se pareshaan hai. {\color{blue}\textit{(Indravardhan is irritated by Madhusudhan's questions)}}} &\multirow{2}{18em}{Indravardhan Madhusudan ke behare pan se pareshaan hai. {\color{blue}\textit{(Indravardhan is tired of Madhusudhan's deafness)}}} &\multirow{2}{18em}{Indravardhan Madhusudan se pareshaan hai. {\color{blue}\textit{(Indravardhan is tired of Madhusudhan)}}} \\
INDRAVARDHAN: Aap ko ka lena dena, panchaanyati laal! {\color{blue}\textit{(What does it have to do with you, Mr. Poke-a-nose?)}} & & & \\ \midrule

MONISHA: Say hello to Tommy the dog. {\color{blue}\textit{(Say hello to Tommy the dog.)}} &\multirow{2}{18em}{Maya monisha ko tana marti hai kyunki usne apne kutte ka naam tommy the dog rakha hai. {\color{blue}\textit{(Maya taunts Monisha on naming her dog Tommy the dog.)}}} &\multirow{2}{18em}{Maya kehti hai ki uske kutte ka naam tommy the dog rakha hai. {\color{blue}\textit{(Maya says that her dog's name is Tommy the dog.)}}} &\multirow{2}{18em}{Maya taunts monisha kyunki usne apne kutte ka naam tommy the dog rakha hai. {\color{blue}\textit{(Maya taunts Monisha that she has named her dog Tommy the dog.)}}} \\
MAYA: Tumne iss kutte ka naam Tommy the dog rakha? {\color{blue}\textit{(Did you name your dog Tommy the dog?)}} & & & \\
\bottomrule
\end{tabular}}
\caption{Actual and generated explanations for sample dialogues from test set. The last utterance is the sarcastic utterance for each dialogue.}
\label{tab:quality}
\vspace{-6mm}
\end{table*}
\subsection{Ablation Study}
To highlight the importance of all modules in consideration, we perform extensive ablation studies on the \datasetSED\ dataset. Table \ref{tab:ablation} shows the results when we add the different proposed modules to our system sequentially. The first row highlights the BART model's results for the text modality which results in a ROUGE-2 of $11.91\%$. As illustrated, the use of naive trimodal concatenation ($T \oplus A \oplus V$) of text, audio, and video representations produces a noisy fusion resulting in decreased performance ($-10.2\%$ ROUGE-2). Next, we try with the standard dot-product attention, which, being a comparatively smarter way of multimodal fusion, results in a slightly improved performance over the text-only modality ($+2\%$ ROUGE-2). Further, adding the multimodal context-aware attention module (MCA2) and replacing standard dot-product attention, produces a further performance boost by $\sim1\%$ across all metrics, signifying the importance of the intelligent fusion that the MCA2 module provides us. The performance is increased even more when the GIF module is introduced to compute the final multimodal vector representation ($+4\%$ ROUGE-2), signifying the positive effect gated fusion has on efficient multimodal representations.
Next, we incorporate pronunciation embeddings (PE) into the model and observe another performance boost across majority metrics ($\sim1\%$), suggesting that we can obtain better code-mixed representations by reducing the spelling ambiguities. Finally, our entire model with modality spotlight included produces the best performance, verifying the necessary use of each module discussed.

\subsection{Result Analysis}
\subsubsection{Quantitative Analysis.}  \model\ is evaluated on its ability to capture sarcasm source and target in the generated explanations. We compare \model\ with mBART, BART, and \textsc{MAF}. Table \ref{tab:source-target} shows that BART performs better than mBART for both source and target detection. The inclusion of multimodal signals, even without pronunciation embeddings and modality spotlight, improves the source identification performance by $\sim14\%$. \model\ is able to detect the sarcasm source most efficiently, resulting in an improvement of $\sim4\%$ over the next best result. Consequently, we can relate the presence of multimodal capabilities to capture speaker-specific peculiarities more efficiently, resulting in better source/target identification.

\begin{table}[!htp]\centering
\resizebox{0.7\columnwidth}{!}{
\begin{tabular}{l|ccc|cc}\toprule
& \textbf{mBART} & \textbf{BART} & \textbf{\textsc{MAF}} & \textbf{\model} \\\midrule
\textbf{Source} &75 &77.23 & \textbf{91.07} &90.17 \\
\textbf{Target} &45.33 &52.67 &46.42 & \textbf{56.69} \\
\bottomrule
\end{tabular}}
\caption{Accuracy for the sarcasm source and target for BART-based systems.}
\label{tab:source-target}
\vspace{-5mm}
\end{table}

\subsubsection{Qualitative Analysis.} We sample a few dialogues from the test set of \datasetSED\ and show their generated explanations by \model\ and the best baseline, \textsc{MAF} along with the ground-truth explanations in Table \ref{tab:quality}. We show one of the many instances where our model generates the correct explanation for the given sarcastic instance in the first row. On the other hand, a lemon-picked instance is shown in the second row, where our model fails to generate the proper description for the explanation. The last row, highlights a case where the generated explanation is not syntactically similar to the ground-truth explanation but resembles it semantically. To evaluate the semantic similarity properly, we perform a human evaluation as explained in the following section.

\subsubsection{Human Evaluation.} We sample a total of $25$ random instances from the test set and ask $20$ human evaluators\footnote{The evaluators are fluent in English and their age ranges in 25-30 years.} to evaluate the generated explanations (on a scale of $1$ to $5$) on the following basis:
\begin{itemize}
    \item \textbf{Coherence:} Checks the generated explanation for correct structure and grammar.
    \item \textbf{On topic:} Measures the extent to which the generated explanation revolves around the dialogue topic.
    \item \textbf{Capturing sarcasm:} Estimates the level of emitted sarcasm being captured in the generated output.
\end{itemize}

We show the average score for the human evaluation parameters in Table \ref{tab:human}. As illustrated, the proposed \model\ model exhibits more coherent, on topic, and sarcasm related explanations. However, there is still a scope for improvement, which can be taken up as future work.

\begin{table}[!htp]\centering
\resizebox{0.8\columnwidth}{!}{
\begin{tabular}{l|cccc}\toprule
&\textbf{Coherency} &\textbf{On topic} &\textbf{Capturing sarcasm} \\\midrule
\textbf{mBART} &2.57 &2.66 &2.15 \\
\textbf{BART} &2.73 &2.56 &2.18 \\
\textbf{\textsc{MAF}} &3.03 &3.11 &2.77 \\\midrule
\textbf{\model} & \textbf{3.96} & \textbf{3.27} & \textbf{3.10} \\
\bottomrule
\end{tabular}}
\caption{Human evaluation statistics -- comparing different models.}
\label{tab:human}
\vspace{-6mm}
\end{table}

\section{Understanding Affects with Explanation}

We study three understanding tasks in dialogues -- sarcasm detection, humour identification, and emotion recognition using \datasetS, \datasetH, and \datasetE, respectively.
A trained SED system is used to obtain the explanations for all the instances present in these datasets. We show the qualitative analysis of the generated explanations by \model\ in the supplementary.
To verify our hypothesis that sarcasm explanation helps affect understanding, we perform experiments with and without explanations, as explained in the subsequent sections.

\subsubsection{Sarcasm Detection.}
\label{sec:sarcasm_detection}
We take a base RoBERTa model \cite{liu2019roberta} and perform the task of sarcasm detection over \datasetS. The experimentation is performed using three setups as described below:
\begin{enumerate}
    \item When we do not provide any utterance explanation to the input dialogue.
    \item When we provide utterances appended with their generated explanation at the training time. Plain dialogues are given at the testing time in this case.
    \item When dialogue instances are appended with their corresponding explanations during train and test time.
\end{enumerate}
Table \ref{tab:affect} illustrates the results we obtain for all the settings for \model\ and the best baseline, \textsc{MAF}. As can be seen, RoBERTa obtains $62\%$ F1 score when we do not use any explanations. 
However, with the use of the generated explanations by \model\ during the train time along with the input dialogues, we obtain an improvement of $6\%$ F1-score.
On the other hand, the best performance is achieved by the last case, where the input instances are appended with their corresponding explanations both at the train and test time, with an increase of $8\%$ F1-score. Consistent to the results obtained by \model's generation, \textsc{MAF} also reports an improved performance over no explanation model. However, the improvement shown by \textsc{MAF} is not at par with the improvement obtained by \model.
These results directly support our hypothesis that utterance explanations can assist an efficient detection of sarcasm in the input instances.


\begin{table}[!htp]\centering
\resizebox{\columnwidth}{!}{
\begin{tabular}{l|cc|rrrr|rrrr|rrrr}\toprule
\multirow{2}{*}{\textbf{Model}} &\multicolumn{2}{c|}{\textbf{Use of Explanation}} &\multicolumn{4}{c|}{\textbf{Sarcasm}} &\multicolumn{4}{c|}{\textbf{Humor}} &\multicolumn{3}{c}{\textbf{Emotion}} \\\cmidrule{2-14}
&\textbf{Train} &\textbf{Test} &\textbf{P} &\textbf{R} &\textbf{F1} &\textbf{Acc} &\textbf{P} &\textbf{R} &\textbf{F1} &\textbf{Acc} &\textbf{P} &\textbf{R} &\textbf{F1} \\\midrule
\textbf{None} &\textbf{0} &\textbf{0} &0.57 &0.68 &0.62 &0.57 &0.69 &0.78 &0.73 &0.87 &0.8 &0.78 &0.78 \\\midrule
\multirow{2}{*}{\textbf{\textsc{MAF}}} &\textbf{1} &\textbf{0} &0.58 &0.73 &0.65 &0.6 &0.57 & \textbf{0.87} &0.69 &0.81 &0.78 &0.78 &0.78 \\
&\textbf{1} &\textbf{1} &0.66 &0.77 &0.71 &0.68 &0.73 &0.71 &0.72 &0.87 &0.78 & \textbf{0.81} &0.79 \\\midrule
\multirow{3}{*}{\textbf{\model}} & \textbf{1} &\textbf{0} &0.65 &0.71 &0.68 &0.66 & \textbf{0.84} &0.63 &0.72 & \textbf{0.89} &0.79 &0.78 &0.78 \\
&\textbf{1} & \textbf{1} & \textbf{0.70} & \textbf{0.83} & \textbf{0.76} & \textbf{0.73} & 0.72 &0.77 & \textbf{0.75} &0.88 & \textbf{0.81} &0.80 & \textbf{0.80} \\
\bottomrule
\end{tabular}}
\caption{Experimental results on RoBERTa base when explanations generated by \model and MAF are used for completing the respective tasks. The first row indicates the performance without explanation. }
\label{tab:affect}
\vspace{-5mm}
\end{table}

\subsubsection{Humour Identification.}
Another RoBERTa base is used to perform humour identification on \datasetH. As for sarcasm detection, humour identification is also evaluated for the three setups described in the previous section. Table \ref{tab:affect} illustrates the results obtained for the described setups. When no explanations are used during the training or testing time, we get an F1-score of $73\%$. This score is comparable to the performance we get when input instances are appended with their corresponding explanations generated by \model\ at the training time. This performance is boosted by $3\%$ when the explanations are provided at the train/test time. However, it is important to note that the explanations generated by the \textsc{MAF} model resulted in a slightly decreased performance indicating the superiority of \model.

\subsubsection{Emotion Recognition.}
Table \ref{tab:affect} illustrates the results obtained for the task of emotion recognition on \datasetE. We see the same value for the weighted F1 when we add explanations during the training phase of the system for both \textsc{MAF} and \model. However, when explanations assist both the training and testing phase, we observe an increase of $2\%$ in the weighted F1 score for \model and $1\%$ increase for \textsc{MAF}, indicating the positive effect explanations deliver for emotion recognition. 
Performance analysis for sarcastic and non-sarcastic instances can be found in the supplementary.

\subsection{Error Analysis}
\subsubsection{Quantitative.} To capture the improvement exhibited by explanations in affect understanding, we show 
the confusion matrices emitted by the understanding models with and without using explanations. Table \ref{tab:affect_confusion} illustrates these matrices --  and as can be seen, the methods with explanation obtains higher true positive rate with a decreased false positive and false negative rates for majority of the classes among sarcasm, humour, and emotion labels.

\begin{table}[!htp]
\centering

\subfloat[\label{tab:affect_analysisH}Sarcasm detection on \datasetS.]
{
\small
\resizebox{0.378\columnwidth}{!}{%
\begin{tabular}{l|ccc}\toprule
\textbf{} & \textbf{NS} & \textbf{S} \\\midrule
\textbf{NS} & \cellcolor{LightGreen} {\color{blue}137}/100 & {\color{blue}81}/117 \\
\textbf{S} & {\color{blue}39}/70 & \cellcolor{LightGreen} {\color{blue}185}/153 \\
\bottomrule
\end{tabular}%
}}
\hspace{1em}
\subfloat[\label{tab:affect_analysisS}{Humour identification on \datasetH.}]
{
\small
\resizebox{0.35\columnwidth}{!}{%
\begin{tabular}{l|ccc}\toprule
\textbf{} & \textbf{NH} & \textbf{H} \\\midrule
\textbf{NH} & \cellcolor{LightGreen} {\color{blue}335}/330 & {\color{blue}32}/37 \\
\textbf{H} & {\color{blue}24}/23 & \cellcolor{LightRed} {\color{blue}82}/83 \\
\bottomrule
\end{tabular}%
}}\\
\subfloat[\label{tab:affect_analysisE}Emotion recognition on \datasetE.]
{
\resizebox{0.7\columnwidth}{!}{%
\begin{tabular}{l|ccccc}\toprule
\textbf{} &\textbf{Neutral} &\textbf{Sadness} &\textbf{Joy} &\textbf{Anger} \\\midrule
\textbf{Neutral} & \cellcolor{LightGreen} {\color{blue}148}/137 & {\color{blue}13}/23 & {\color{blue}18}/19 & {\color{blue}16}/16 \\
\textbf{Sadness} & {\color{blue}5}/2 & \cellcolor{LightRed} {\color{blue}62}/66 & {\color{blue}3}/2 & {\color{blue}0}/0 \\
\textbf{Joy} & {\color{blue}7}/5 & {\color{blue}10}/9 & \cellcolor{LightRed} {\color{blue}120}/124 & {\color{blue}4}/3 \\
\textbf{Anger} & {\color{blue}0}/9 & {\color{blue}0}/1 & {\color{blue}8}/9 & \cellcolor{LightGreen} {\color{blue}50}/48 \\
\bottomrule
\end{tabular}%
}}
\caption{Confusion matrix of the systems {\color{blue}with} and without explanations.}
\label{tab:affect_confusion}
\vspace{-4mm}
\end{table}


\subsubsection{Qualitative.} While quantitative results confirm that explanations assist in identifying affects efficiently, qualitative analysis can further corroborate this hypothesis. Table \ref{tab:affect_quality} shows one instance from the test set where the presence of explanation helps for all affective tasks in question. More such examples can be found in the supplementary.

\begin{table}[!htp]\centering
\resizebox{\columnwidth}{!}{
\begin{tabular}{l|p{10em}p{10em}p{10em}r}\toprule

\multirow{3}{*}{\textbf{Dialogue}} &\multicolumn{3}{p{30em}}{\textbf{MAYA:} And this time I thought lets have a theme party! {\color{blue}\textit{(And this time I thought lets have a theme party!)}}} \\
&\multicolumn{3}{p{30em}}{\textbf{MONISHA:} Animals! Hum log sab animals banenge! {\color{blue}\textit{(Animals! Let us all be animals this time!)}} } \\
&\multicolumn{3}{p{30em}}{\textbf{MAYA:} Mai hiran, Sahil horse, and Monisha chhipakalee! {\color{blue}\textit{(I'll be a deer, Sahil a horse, and Monisha can be a lizard!)}}} \\\midrule

\textbf{Exp} &\multicolumn{3}{p{30em}}{Maya Monisha ko animal keh ke taunt maarti hai. {\color{blue}\textit{(Maya taunts Monisha by calling her an animal)}}} \\\midrule

\textbf{} &\multicolumn{1}{m{10em}|}{\centering \textbf{Sarcasm}} & \multicolumn{1}{m{10em}|}{\centering \textbf{Humour}} & \multicolumn{1}{m{10em}}{\centering \textbf{Emotion}} \\\midrule

\textbf{GT} & \multicolumn{1}{c|}{1} & \multicolumn{1}{c|}{0} & \multicolumn{1}{c}{Anger} \\
\textbf{w/o Exp} & \multicolumn{1}{c|}{\cellcolor{LightRed} 0} & \multicolumn{1}{c|}{\cellcolor{LightRed} 1} & \multicolumn{1}{c}{\cellcolor{LightRed} Neutral} \\
\textbf{w Exp} & \multicolumn{1}{c|}{\cellcolor{LightGreen} 1} & \multicolumn{1}{c|}{\cellcolor{LightGreen} 0} & \multicolumn{1}{c}{\cellcolor{LightGreen} Anger} \\
\bottomrule
\end{tabular}}
\caption{True and predicted labels for the three affect tasks with and without using \model's explanation.}
\label{tab:affect_quality}
\vspace{-5mm}
\end{table}

\section{Conclusion}
The inability of existing systems to understand sarcasm results in a performance gap for various affect understanding tasks like emotion recognition, humour identification, and sarcasm detection. To mitigate this issue, we proposed \model\ to explore the task of Sarcasm Explanation in Dialogues (SED). \model\ takes multimodal code-mixed sarcastic conversation instances as input and results in a natural language explanation describing the sarcasm present in it. We further explored the effect of the generated explanations on three dialogue-based affect understanding tasks -- sarcasm detection, humour identification, and emotion recognition. We observed that explanations improved the performance of all three tasks, thus verifying our hypothesis.

\section*{Acknowledgement} The authors acknowledge the support of the ihub-Anubhuti-iiitd Foundation, set up under the NM-ICPS scheme
of the DST, and CAI-IIITD.


\bibliography{aaai23}

\section{Appendix}
\begin{table*}[t]\centering
\resizebox{\textwidth}{!}{
\subfloat[For \datasetSED]{
\begin{tabular}{l|p{20em}|r|p{20em}|p{20em}r}\toprule
\textbf{Context Speakers} &\textbf{Context Utterances} &\textbf{Target Speaker} &\textbf{Target Sarcastic Utterance} &\textbf{Explanation} \\\midrule
INDRAVARDHAN &Accha suno Monisha tumhaare ghar mein been ya aisa kuuch hain? {\color{blue}\textit{(Listen Monisha, do you have a flute or something similar?)}} &MAYA &Kaise hogi? Monisha aapne ghar pe dustbin mushkil se rakhti hain to snake charmer waali been kaha se rakhegi? {\color{blue}\textit{(How will it be there? Monisha hardly keeps a dustbin in her home so how will she has a snake charmer’s flute?)}} &Maya Monisha ko tana marti hai safai ka dhyan na rakhne ke liye. {\color{blue}\textit{(Maya taunts Monisha for not keeping a check of cleanliness)}} \\\midrule

SAHIL &Ab tumne ghar ki itni saaf safai ki hai and secondly us Karan Verma ke liye pasta, lasagne, caramel custard banaya. {\color{blue}\textit{(Now you have cleaned the house so much and secondly made pasta, lasagne, caramel custard for that Karan Verma.)}} &\multirow{2}{*}{SAHIL} &\multirow{2}{20em}{Walnut brownie, matlab wo khane wali? {\color{blue}\textit{(You mean edible walnut brownie?)}}} &\multirow{2}{20em}{Sahil monisha ki cooking ka mazak udata hai. {\color{blue}\textit{(Sahil makes fun of Monisha’s cooking.)}}} \\
MONISHA &Walnut brownie bhi. {\color{blue}\textit{(And walnut brownie too.)}} & & & \\
\bottomrule
\end{tabular}
\label{tab:instance_sed}
}}\\
\vspace{5mm}

\resizebox{\textwidth}{!}{
\subfloat[For \datasetS, \datasetH, and \datasetE]{
\begin{tabular}{l|p{20em}|r|p{20em}|c|c|cr}\toprule
Context Speakers &Context Utterances &Target Speaker &Target Utterance & \datasetS & \datasetH & \datasetE \\\midrule

MONISHA &Dukan se yaad aya, mummy ji wahan pe Southhall ya Wembley ki kisi dukan se please mere liye chandi ka mangalsutra le aaiega na {\color{blue}\textit{(Talking about shops, mom please get me a silver necklace from any shop from Southhall or Wembley.)}} &\multirow{2}{*}{MONISHA} &\multirow{2}{20em}{Haan wahan pe kali mani ke neeche Big Ben ki pendant wala mangalsutra milta hai. {\color{blue}\textit{(Yes, we can get a necklace of black beads from there.)}}} &\multirow{2}{*}{0} &\multirow{2}{*}{1} &\multirow{2}{*}{Joy} \\

SAHIL &Mangalsutra London se? {\color{blue}\textit{(You want a necklace from London?)}} & & & & & \\\midrule

ROSESH &Momma mujhe bohot achi lagti hai. {\color{blue}\textit{(I like momma very much)}} &\multirow{3}{*}{INDRAVARDHAN} &\multirow{3}{20em}{Rakshas! {\color{blue}\textit{(Monster!)}}} &\multirow{3}{*}{1} &\multirow{3}{*}{0} &\multirow{3}{*}{Anger} \\

INDRAVARDHAN &I know that. Momma pari hai pari! {\color{blue}\textit{(I know that. Your mother is like a fairy.)}} & & & & & \\
ROSESH &Aur me? {\color{blue}\textit{(And me?)}} & & & & & \\
\midrule
\end{tabular}
\label{tab:instance_she}
}}
\caption{Sample instances for \datasetSED, \datasetS, \datasetH, and \datasetE}
\label{tab:instance}
\end{table*}

\subsection{Dataset}
\subsubsection{Curation of \datasetS, \datasetH, and \datasetE}
All affects such as sarcasm, humour, and emotions work in tandem to unveil the true connotation of an utterance in a dialogue. Thereby, a comprehensive justification for one component should help in improving the understanding of the others. In this work, we deal with the task of Sarcasm Explanation in Dialogues (SED) \citep{kumar2022did} where the aim is to generate a natural language explanation for a given multimodal sarcastic conversation. We propose \model, a deep neural architecture with the aim to solve SED. To exhibit the influence of sarcasm on other affective components, like humour and emotions, we leverage the generated explanation by \model\ and augment the input instances for affective tasks with them. We experiment with three affect understanding tasks in dialogues -- sarcasm detection, humour identification, and emotion recognition. To perform appropriate experimentation for the tasks with and without explanations, we require supporting datasets. Consequently, we curate \datasetS, \datasetH, and \datasetE for the task of sarcasm detection, humour identification, and emotion recognition, respectively, from the \datasetSED\ dataset \citep{kumar2022did}. 

\begin{figure}[h!]
    \centering
    \includegraphics[width=\columnwidth]{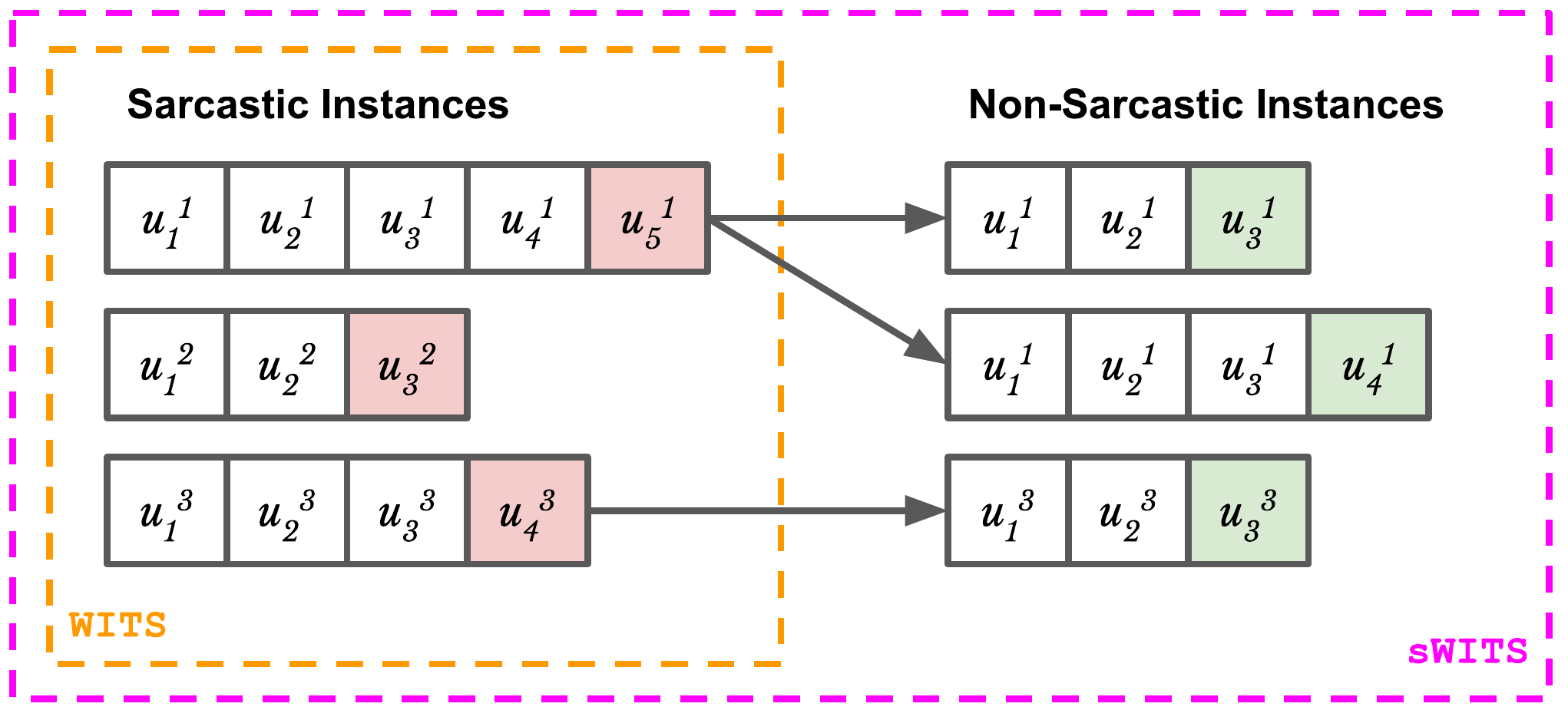}
    \caption{Construction of \datasetS\ from \datasetSED.}
    \label{fig:instance_creation}
\end{figure}

\paragraph{\datasetS:} The parent dataset, \datasetSED\, contains sarcastic instances along with their explanations. Each instance of \datasetSED\ contains a sequence of utterances where the last utterance is sarcastic. However, for the sarcasm detection task, we need both sarcastic as well as non-sarcastic instances. To create the non-sarcastic instances, we randomly sample utterances from the context of the instances present in \datasetSED. Figure \ref{fig:instance_creation} illustrates the process of creating \datasetS\ from \datasetSED.

\paragraph{\datasetH:} To gauge the effect of sarcasm explanation on humour identification, we will need instances with humour labels. As a result, we explore the mapping existing between \datasetSED\ and the \textsc{MaSaC} dataset \citep{9442359}. \textsc{MaSaC} is a multiparty, multimodal, code-mixed dialogue dataset collected from an Indian TV series. It contains binary markers identifying the presence of sarcasm and humour in all utterances. The \datasetSED\ dataset is an extended version of the \textsc{MaSaC} dataset where all the sarcastic utterances are appended with their corresponding natural language explanation. Ergo, we map the humour labels from \textsc{MaSaC} to the instances present in \datasetS\ and get \datasetH.

\paragraph{\datasetE:} Sarcasm significantly affects the emitted emotion of an utterance. Wherefore, we hypothesize that emotion recognition in conversation can be improved in the presence of utterance explanations. Therefore, we need emotion labels for the instances present in \datasetS\ to create \datasetE. We annotate the instances for emotion labels following the Ekman \cite{Ekman92anargument} emotion scheme. Out of the seven possible emotion labels, namely \textit{anger}, \textit{fear}, \textit{disgust}, \textit{sadness}, \textit{joy}, and \textit{surprise}, \textit{neutral}, we were able to identify four for our set of instances -- \textit{anger}, \textit{sadness}, \textit{joy}, and \textit{neutral}. Three annotators ($ABC$) were involved and achieved Krippendorff's Alpha \cite{krippendorff2011computing} inter-annotator scores as $\alpha_{AB} = 0.84$, $\alpha_{BC} = 0.88$, and $\alpha_{AC} = 0.86$ giving an average score of $0.86$. We show couple of instances from all the discussed datasets in Table \ref{tab:instance}.

\subsection{Quality of Generated Explanations}
Our model, \model, being taught the Sarcasm Explanation in Dialogues (SED) task, is supplied only with sarcastic instances during training. Consequently, it is able to efficiently generate explanations for unseen sarcastic instances. However, when supplied with non-sarcastic instances, as is done to generate \datasetS, \datasetH, and \datasetE, one can argue that the model might not work. Nevertheless, \model\ is able to return appropriate explanations for the majority of the input. Although, there are some explanations generated containing an undertone of sarcasm even for non-sarcastic input instances. Table \ref{tab:exp_eg} illustrates a few sample explanations generated for sarcastic and non-sarcastic inputs. The first two instances are sarcastic in nature and the corresponding explanations try to reveal this irony. While the first explanation seems relevant to the input dialogue, the explanation in the second instance is an example of incorrect generation. The last instance in Table \ref{tab:exp_eg} is non-sarcastic and as shown, the explanation generated seems correct for the dialogue. However, improving upon the quality of explanations, especially for non-sarcastic inputs, can be an interesting future direction of research.

\begin{table}[!htp]\centering
\resizebox{\columnwidth}{!}{
\begin{tabular}{p{20em}|p{10em}r}\toprule
\textbf{Dialogue} &\textbf{Generated Explanation} \\\midrule
ROSESH : Sahil bhai there is a pattern {\color{blue}\textit{(Brother Sahil, there is a pattern.)}} & \multirow{6}{10em}{Sahil Roshesh ki acting ka mazak udata h {\color{blue}\textit{(Sahil makes fun of Rosesh's acting)}}} \\
SAHIL : Kaisa pattern? {\color{blue}\textit{(What pattern?)}} & \\
ROSESH : Pahli bar Yamraj Chhole Bhature Dekhkar bagaya tha is bar rasgulle {\color{blue}\textit{(First time Yamraj ran away after seeing Chhole Bhature (food) and this time Rasgulle (sweet))}} & \\
MONISHA : Matlab? {\color{blue}\textit{(Meaning?)}} & \\
ROSESH : Matlab Yamraj tra troo tree. {\color{blue}\textit{(Means Yamraj tra troo tree (mimicking elephant))}} & \\
SAHIL : Are vah iska kya matlab? {\color{blue}\textit{(Very nice. What does it mean?)}} {\color{red}[Sarcastic]} & \\\midrule
MR. SINDOLIN : Okay... {\color{blue}\textit{(Okay...)}} &\multirow{4}{10em}{Indravardhan Siddharth ko mystic keh ke uska majaak udaata hai. {\color{blue}\textit{(Indravardhan makes fun of Siddharth by calling him a mystic.)}}} \\
SAHIL : Ocean of the unknown... {\color{blue}\textit{(Ocean of the unknown...)}} & \\
MAYA : Mountains of mystic... {\color{blue}\textit{(Mountains of mystic...)}} & \\
MR. SINDOLIN : Ek minute, ek minute, mai jaau? {\color{blue}\textit{(One minute, one minute, should I leave?)}} {\color{red}[Sarcastic]} & \\\midrule
MONISHA : la la la la la la la {\color{blue}\textit{(la la la.. (singing))}} &\multirow{2}{10em}{Maya ko lagta hai ki log log pagal ho gaya hai. {\color{blue}\textit{(Maya thinks these people have gone mad.)}}} \\
MAYA : Tum log pagal ho gye ho? {\color{blue}\textit{(Have you guys gone mad?)}} {\color{teal}[Non Sarcastic]} & \\
&\\
\bottomrule
\end{tabular}}
\caption{Examples of explanations generated by \model\ for sarcastic and non-sarcastic instances.}
\label{tab:exp_eg}
\vspace{-5mm}
\end{table}

\subsection{Error Analysis for Affect Understanding}
\subsubsection{Quantitative}
Since \model\ is trained on the task of SED, there is a need to investigate its performance separately for sarcastic and non-sarcastic instances. We examine the obtained performance for sarcastic and non-sarcastic instances separately with and without using the generated dialogue explanation and produce those results in Table \ref{tab:affect_sar_nsar} for the task of humour identification and emotion recognition. We observe that while the performance of sarcastic humorous instances decreased after the addition of generated explanations; the performance of non-sarcastic humorous dialogues increased by $8\%$ when explanations are appended with the input. On the contrary, for the sarcastic emotion instances, the performance is enhanced in the presence of explanations while it remains comparable for non-sarcastic instances.

\begin{table}[!htp]\centering
\resizebox{\columnwidth}{!}{
\begin{tabular}{lr|rrr|rrrr}\toprule
\multicolumn{2}{c}{\multirow{2}{*}{\textbf{}}} &\multicolumn{3}{|c|}{\textbf{Humour}} &\multicolumn{3}{c}{\textbf{Emotion}} \\\cmidrule{3-8}
& &\textbf{P} &\textbf{R} &\textbf{F1} &\textbf{P} &\textbf{R} &\textbf{F1} \\\midrule
\multirow{2}{*}{\textbf{Sarcastic}} &\textbf{w/o exp} &0.74 &\textbf{0.85} &0.79 &0.81 &0.81 &0.81 \\
&\textbf{w exp} &0.71 &0.73 &0.72 &\textbf{0.83} &\textbf{0.82} &\textbf{0.82} \\\midrule
\multirow{2}{*}{\textbf{Non-sarcastic}} &\textbf{w/o exp} &0.75 &0.74 &0.75 &0.78 &0.77 &0.77 \\
&\textbf{w exp} &\textbf{0.83} &0.83 &\textbf{0.83} &0.77 &0.76 &0.76 \\
\bottomrule
\end{tabular}}
\caption{Experimental results on sarcastic and non-sarcastic instances using the RoBERTa base model with and without explanation assistance.}
\label{tab:affect_sar_nsar}
\end{table}

\subsubsection{Qualitative}
Table \ref{tab:affect_quality2} highlights some more sample cases from our test sets where the presence of explanations influence the efficiency of affect classification. While there are cases where the presence of the generated explanations improve the results of the concerned tasks, there are a few cases where the explanation might add confusion for the system. We show these instances in Table \ref{tab:affect_quality2}.

\subsection{Training System}
We mention below the computational framework we use to train our models.
\begin{itemize}
    \item Description of computing infrastructure used
        \begin{itemize}
            \item Linux $64$ Bit
            \item GPU: Tesla-V100 ($32510$ MiB)
        \end{itemize}
    \item Trainable parameter: $326368976$
    \item Average runtime: 180 seconds per epoch
    \item All the results are an average of $3$ runs.
\end{itemize}

\subsubsection{Hyperparameter Tuning}
\begin{table}[!htp]\centering
\resizebox{\columnwidth}{!}{
\begin{tabular}{lr|rr|rr|rrr}\toprule
\multicolumn{2}{c|}{\textbf{General}} &\multicolumn{2}{c|}{\textbf{Visual Transformer}} &\multicolumn{2}{c|}{\textbf{Acoustic Transformer}} &\multicolumn{2}{c}{\textbf{PE Transformer}} \\\midrule
\textbf{Batch size} &4 &\textbf{\# Layers} &4 &\textbf{\# Layers} &4 &\textbf{\# Layers} &4 \\
\textbf{Learning rate} &5e-6 &\textbf{\# Heads} &8 &\textbf{\# Heads} &2 &\textbf{\# Heads} &2 \\
\textbf{Weight decay} &1e-4 &\textbf{Dimension} &2048 &\textbf{Dimension} &154 &\textbf{Dimension} &154 \\
\bottomrule
\end{tabular}}
\caption{Optimal hyperparameters selected for \model.}
\label{tab:hyperparameters}
\end{table}

After careful manual tuning of the hyperparameters we come up with the optimal set of parameters as mentioned in Table \ref{tab:hyperparameters}. After considering batch size ranging from $2$-$8$, we set the batch size to $4$ due to the computational bottleneck. The learning rate was selected as $5e-6$ with a weight decay of $1e-4$ as a lower learning rate resulted in extremely slow training while a higher rate resulted in haphazard learning. Additionally the number of layers and heads for each Transformer encoder used is selected after trying different combinations.



\begin{table}[!htp]\centering
\subfloat[All affect classification improves.]{
\label{tab:affect_quality_best}
\resizebox{\columnwidth}{!}{
\begin{tabular}{l|p{10em}p{10em}p{10em}r}\toprule

\multirow{3}{*}{\textbf{Dialogue}} &\multicolumn{3}{p{30em}}{\textbf{MONISHA:} Dukan se yaad aya mummy ji, wahan pe Southhall ya Wembley ki kisi dukan se please mere liye chandi ka mangalsutra le aayiega na. {\color{blue}\textit{(Talking about shops, mom please get me a silver necklace from any shop from Southhall or Wembley.)}}} \\
&\multicolumn{3}{p{30em}}{\textbf{SAHIL:} Magalsutra, London se? {\color{blue}\textit{(You want a necklace from London?)}}} \\

&\multicolumn{3}{p{30em}}{\textbf{MONISHA:} Haan, wahan pe kali mani ke neeche big ben ke pendant wala mangalsutra milta hai. {\color{blue}\textit{(Yes, we can get a necklace of black beads from there.)}}} \\
&\multicolumn{3}{p{30em}}{\textbf{MAYA:} Monisha me Central London me rehne wali hun, Piccadilly Circus ke paas. Wembley ya southhall jesi middle class jagah me nhi. {\color{blue}\textit{(Monisha, I'll be staying in Central London, near Piccadilly Circus. Not at some middle class place like Wembley or Southhall.)}}} \\\midrule

\textbf{Exp} &\multicolumn{3}{p{30em}}{Maya Monisha ke style ka mazak udati hai. {\color{blue}\textit{(Maya makes fun of Monisha's style.)}}} \\\midrule

\textbf{} &\multicolumn{1}{m{10em}|}{\centering \textbf{Sarcasm}} & \multicolumn{1}{m{10em}|}{\centering \textbf{Humour}} & \multicolumn{1}{m{10em}}{\centering \textbf{Emotion}} \\\midrule

\textbf{GT} & \multicolumn{1}{c|}{1} & \multicolumn{1}{c|}{1} & \multicolumn{1}{c}{Anger} \\
\textbf{w/o Exp} & \multicolumn{1}{c|}{\cellcolor{LightRed} 0} & \multicolumn{1}{c|}{\cellcolor{LightRed} 0} & \multicolumn{1}{c}{\cellcolor{LightRed} Neutral} \\
\textbf{w Exp} & \multicolumn{1}{c|}{\cellcolor{LightGreen} 1} & \multicolumn{1}{c|}{\cellcolor{LightGreen} 1} & \multicolumn{1}{c}{\cellcolor{LightGreen} Anger} \\
\bottomrule
\end{tabular}}}\\

\subfloat[Some affect classification improves.]{
\label{tab:affect_quality_better}
\resizebox{\columnwidth}{!}{
\begin{tabular}{l|p{10em}p{10em}p{10em}r}\toprule

\multirow{3}{*}{\textbf{Dialogue}} &\multicolumn{3}{p{30em}}{\textbf{INDRAVARDHAN:} Wo tum Rosesh se hi puch lena kal, wo lekar aa rha yahan apni dharam patni ko. {\color{blue}\textit{(You can ask this to Rosesh, he is bringing his lawful wife here tomorrow.)}}} \\

&\multicolumn{3}{p{30em}}{\textbf{MAYA:} Indravadan please. Dharam patni is television serial middle class. {\color{blue}\textit{(Indravardhan, please. Lawful wise if television serial middle class.)}} } \\\midrule

\textbf{Exp} &\multicolumn{3}{p{30em}}{Maya ko lagta hai ki dharam patni is television serial middle class. {\color{blue}\textit{(Maya thinks that lawful wife is television serial middle class.)}}} \\\midrule

\textbf{} &\multicolumn{1}{m{10em}|}{\centering \textbf{Sarcasm}} & \multicolumn{1}{m{10em}|}{\centering \textbf{Humour}} & \multicolumn{1}{m{10em}}{\centering \textbf{Emotion}} \\\midrule

\textbf{GT} & \multicolumn{1}{c|}{1} & \multicolumn{1}{c|}{0} & \multicolumn{1}{c}{Neutral} \\
\textbf{w/o Exp} & \multicolumn{1}{c|}{\cellcolor{LightGreen} 1} & \multicolumn{1}{c|}{\cellcolor{LightGreen} 0} & \multicolumn{1}{c}{\cellcolor{LightGreen} Neutral} \\
\textbf{w Exp} & \multicolumn{1}{c|}{\cellcolor{LightGreen} 1} & \multicolumn{1}{c|}{\cellcolor{LightRed} 1} & \multicolumn{1}{c}{\cellcolor{LightRed} Anger} \\
\bottomrule
\end{tabular}}}\\

\subfloat[No affect classification improves.]{
\label{tab:affect_quality_worse}
\resizebox{\columnwidth}{!}{
\begin{tabular}{l|p{10em}p{10em}p{10em}r}\toprule

\multirow{3}{*}{\textbf{Dialogue}} &\multicolumn{3}{p{30em}}{SAHIL: Almaari itni bhi buri nahi hai mom. Not bad... {\color{blue}\textit{(The cupboard is not that bad mom. Not bad...))}}} \\
&\multicolumn{3}{p{30em}}{MAYA: Nah, its not bad... its just atrocious! {\color{blue}\textit{(No, its not bad... its just atrocious!)}} } \\\midrule

\textbf{Exp} &\multicolumn{3}{p{30em}}{Maya Sahil ko uske eating habits par taunt maarti hai. {\color{blue}\textit{(Maya taunts Sahil for his eating habits.)}}} \\\midrule

\textbf{} &\multicolumn{1}{m{10em}|}{\centering \textbf{Sarcasm}} & \multicolumn{1}{m{10em}|}{\centering \textbf{Humour}} & \multicolumn{1}{m{10em}}{\centering \textbf{Emotion}} \\\midrule

\textbf{GT} & \multicolumn{1}{c|}{1} & \multicolumn{1}{c|}{1} & \multicolumn{1}{c}{Anger} \\
\textbf{w/o Exp} & \multicolumn{1}{c|}{\cellcolor{LightGreen} 1} & \multicolumn{1}{c|}{\cellcolor{LightRed} 0} & \multicolumn{1}{c}{\cellcolor{LightGreen} Anger} \\
\textbf{w Exp} & \multicolumn{1}{c|}{\cellcolor{LightRed} 0} & \multicolumn{1}{c|}{\cellcolor{LightRed} 0} & \multicolumn{1}{c}{\cellcolor{LightRed} Sadness} \\
\bottomrule
\end{tabular}}}
\caption{True and predicted labels for the three affect tasks with and without using \model's explanation.}
\label{tab:affect_quality2}
\vspace{-5mm}
\end{table}
\end{document}